\documentclass[runningheads]{llncs}

\usepackage[T1]{fontenc}
\def\doi#1{\href{https://doi.org/\detokenize{#1}}{\url{https://doi.org/\detokenize{#1}}}}
\usepackage{graphicx}
\usepackage{multicol}
\usepackage{amsmath}
\usepackage{bbding}

\usepackage{multirow}
\usepackage{listings}
\begin{document}
%
% \title{An interpretable loan credit evaluation method based on RRL\thanks{Supported by organization x.}}
\title{An Interpretable Loan Credit Evaluation Method Based on Rule Representation Learner}
% AUTHOR INFO BEGIN
\author{Zihao Chen\inst{1} \and
Xiaomeng Wang \textsuperscript{\Envelope}\inst{1}
\and
Yuanjiang Huang\inst{2}
\and
Tao Jia \inst{1}
}
\authorrunning{Z. Chen et al.}
\institute{Southwest University, Chongqing 400700, China\\
\email{chenzihao@email.swu.edu.cn,\\
\{wxm1706,tjia\}@swu.edu.cn }\\
\and
BaiHang Intelligent Data Technology Institute, Chongqing, China\\
\email{yuanjiang.huang@socialcredits.cn}  }
% AUTHOR INFO END
%
\maketitle              % typeset the header of the contribution
\begin{abstract}
The interpretability of model has become one of the obstacles to its wide application in the high-stake fields. The usual way to obtain interpretability is to build a black-box first and then explain it using the post-hoc methods. However, the explanations provided by the post-hoc method are not always reliable. Instead, we design an intrinsically interpretable model based on RRL(Rule Representation Learner) for the Lending Club dataset. Specifically, features can be divided into three categories according to their characteristics of themselves and build three sub-networks respectively, each of which is similar to a neural network with a single hidden layer but can be equivalently converted into a set of rules. During the training, we learned tricks from previous research to effectively train binary weights. Finally, our model is compared with the tree-based model. The results show that our model is much better than the interpretable decision tree in performance and close to other black-box, which is of practical significance to both financial institutions and borrowers. More importantly, our model is used to test the correctness of the explanations generated by the post-hoc method, the results show that the post-hoc method is not always reliable.

\end{abstract}
\section{Introduction}
Credit is a core concept in the financial field, and credit scoring and rating are widely studied problems with a long history\cite{thomas2017credit}. When borrowers apply for loans, banks or institutions are expected to make decisions not only agilely but also precisely. In the massive data environment created by financial technology (FinTech), technologies such as machine learning and data mining have become important technical means for credit evaluation due to their powerful data analysis capabilities.Some research and achievements related have been witnessed\cite{de2018advances,goodell2021artificial,nti2020systematic,ozbayoglu2020deep,zheng2019finbrain}.

However, since the period of statistical learning, the machine learning community has been focusing on improving performance and pursuing the improvement of various performance metrics. This phenomenon is particularly obvious since entering the era of deep learning \cite{lecun2015deep}. \cite{zhang2021survey} points out that with the increasing strength of deep neural networks and their popularity in our lives, there is also growing concerned about their black-box nature. If domain experts can't explain the decision-making process of the model, how can users trust it? Black-box refers to a model that does not understand the internal mechanism or explain the decision-making process. Such a model can not be directly used in domains demanding high transparency, such as financial, medicine, criminal judicature, and other high-stake decisions. It can be said that the lack of interpretability has become one of the obstacles to a wide application in these fields.

Therefore, interpretable machine learning has attracted more and more attention in the past years. The usual way to obtain interpretability is to build a black-box model first and then explain its behavior using the post-hoc methods. However, many researchers are skeptical about the reliability of this direction. Considering interpretability before modeling is encouraged \cite{rudin2019stop}. Some works of interpretable machine learning tend to propose a general model or algorithm \cite{ribeiro2016should,ribeiro2018anchors,lundberg2017unified}, but it is unknown whether that model meets the requirements in the actual application scenarios. In application scenarios with high requirements for interpretation, it is unrealistic to seek a one-for-all model or interpretation technology, because different application scenarios need different interpretations. For example, in image classification applications\cite{NEURIPS2019_adf7ee2d}, a partial area of the image is enough for interpretation of the classification results, not specific to each feature (pixel). For healthcare and criminal justice, the traditional scorecard is more acceptable for practitioners \cite{rudin2018optimized}, but for credit evaluation, while the rule-based interpretation may be more user-friendly. RRL (Rule Representation Learner) \cite{wang2021scalable} provides a rule representation and learning framework, which has performance advantages over traditional decision tree methods. So we design a credit evaluation model based on RRL. The main contributions of the paper are as follows:

\begin{itemize}
		\item[$\bullet$] Focusing on the credit evaluation, we designed an interpretable loan credit evaluation model based on RRL. Our model can naturally extract accurate global and local explanations without using the post-hoc methods. It has practical significance for both financial institutions and borrowers.
		\item[$\bullet$] Our model is compared and validated with other tree-based models on the Lending Club dataset. The results show that our performance is far better than that of interpretable CART, and is close to the black box, like Random Forest, XGBoost, and LightGBM.
		\item[$\bullet$] An experiment is designed to verify the correctness of the post-hoc interpretation methods based on our model, results show why it is not suitable to use the post-hoc methods in high-stake decision-making scenarios.
\end{itemize}

The rest of this paper is organized as follows. Section 2 introduces the related work and development of interpretable machine learning and credit risk evaluation. Section 3 presents the structure and training process of our model. Section 4 shows the experimental results compared with the tree-based model, and also illustrates how our model provides local and global explanations. An experiment to verify the correctness of the post-hoc interpretation method is also designed. Section 5 concludes the paper and covers future work.

\section{Related Work}

\subsection{Interpretable Machine Learning}
Interpretable machine learning has attracted much attention in recent years\cite{2020Interpretable,2021Interpretable}. As suggested by  \cite{molnar2020interpretable}, it can be simply divided into post-hoc methods and ante-hoc (or intrinsically interpretable design, white-box). The post-hoc methods do not check the internal structure or parameters of the original model. They attempt to provide explanations of behavior from the trained model (the black box). Therefore, they are available to all kinds of models, and it has been widely studied in the past few years. Such representative research includes LIME\cite{ribeiro2016should}, Anchor\cite{ribeiro2018anchors}, SHAP\cite{lundberg2017unified,lundberg2018consistent}, etc. For example, LIME can get a locally linear surrogate model as an explanation for an individual sample, Anchor provides sufficient explanations for the decision-making of the model. It has to be said that the emergence of post-hoc methods improves users' understanding in most cases. However, there is no such thing as a free lunch. Some researchers raised concerns \cite{rudin2019stop,laugel2019dangers,slack2020fooling}. If explanations provided by the post-hoc methods can faithfully reflect the black-box model, is it necessary for the black box to exist? If not, why should we believe the explanations it provides.

The post-hoc methods apply to fields with low safety requirements, Such as it will not bring too much loss even if the interpretation is misleading. But for other high-stakes fields, post-hoc has not seemed a sensible choice. According to the ante-hoc, related studies can be divided into two categories.

Optimization of traditional statistical learning models and mathematical programming, including improving the training efficiency of the interpretable algorithm and regularization constraints on the complexity. For examples: \cite{hu2019optimal,lin2020generalized} focused on the research of decision tree and made efforts to acquire a sparse but also accurate decision tree according to a custom objective function. \cite{ustun2019learning} solved a mixed integer nonlinear program problem in an acceptable time and get an integer scorecard on the premise of a performance guarantee.

Another way is to design a model that meets interpretability requirements in specific application scenarios: \cite{kim2020stepwise} decomposes a regression problem, in which LSTM is responsible for ensuring performance and linear regression is responsible for explaining. \cite{chen2022holistic} built a two-layer additive risk model in Explainable Machine Learning Challenge organized by FICO, and it is comparable to black-box models in performance. \cite{agarwal2021neural} presented neural additive models which combine some of the expressivity of DNNs with the inherent interpretability of generalized additive models. \cite{wang2021scalable} tried to extract conjunction or disjunction rules from neural networks for classification tasks. Generally speaking, intrinsically interpretable models is a broad topic, and it is also a mainstream trend of interpretable machine learning. Modelers are required to design exclusive interpretable models for specific application scenarios. Our model also belongs to this category.

\subsection{Credit Risk Evaluation}
Credit risk is one of the three risks defined in the Basel Accord \cite{thomas2017credit}. Personal credit risk evaluation is an essential aspect of financial risk management. With the rapid development of digital finance, government, banks, financial institutions, and FinTech companies have accumulated a large amount of data, providing a solid data foundation for credit risk modeling. In the loan application scenario, a functional model can predict the solvency and willingness of users according to features collected, provide decision support for transactions, facilitate applicants and help financial institutions avoid risks, which is conducive to promoting sound development of the financial market.

Personal credit risk evaluation is a typical classification task suitable for modeling by machine learning. In the past decade, a large number of machine learning methods have been applied to credit scoring or evaluation \cite{baesens2003benchmarking,lessmann2015benchmarking,moscato2021benchmark}. Among a huge amount of algorithms, logistic regression is still widely used today because of its solid statistical foundation and strong interpretability. Deep learning, and ensemble learning are usually superior in performance, but they are difficult to be widely used in credit risk scenarios for many reasons \cite{gunnarsson2021deep}. Interpretability is one of the major obstacles. Therefore, applying interpretable machine learning techniques to the credit evaluation during loan application is natural.

\section{Model}

If the loan application of a customer is rejected, the financial institution needs to clarify the reason. For example, it is rejected for ``The total assets are less than 10000 and the monthly income is less than 5000''. This requires that the causal relationship between the inputs and outputs in the credit evaluation model is clear. Although the decision tree model is easy to do, it has shortcomings in heuristic training and performance. Instead, RRL (Rule-based Representation Learner)\cite{wang2021scalable} is applied to construct a interpretable credit evaluation model in the paper. The key to model construction includes logical rule determination, feature selection and binarization, and model training.

\subsection{Rule Representation}

For user $i$, the rejection of his loan application is determined by a combination of factors. Then, we simply formalize the interpretation as the equation (\ref{rule}).

\begin{equation}\label{rule}
E_{i} = r_{i}^{1} \wedge r_{i}^{2} ... \wedge r_{i}^{n}
\end{equation}

Where $E_{i}$ indicates that this is an interpretation set for rejected customer $i$. $r_{i}^{n}$ indicates the $n$th meta rule for customer $i$. Such as ``The total assets are less than 10000''.

\begin{figure}
\centering
\includegraphics[width=0.6\textwidth]{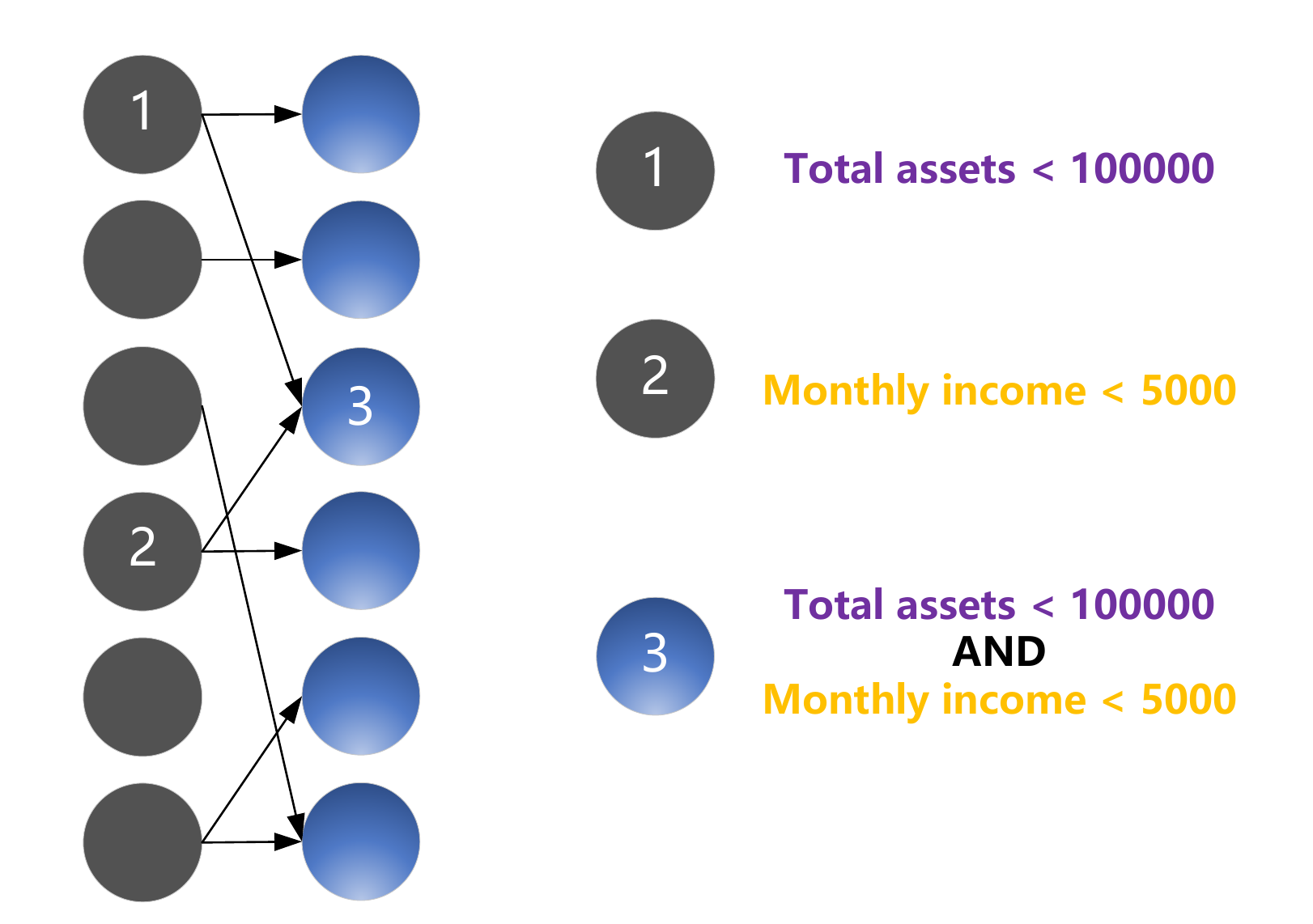}
\caption{An example: represents conjunction rules by a neural network.} \label{fig1}
\end{figure}

A neural network (Fig.~\ref{fig1}) and a conjunction function\cite{payani2019learning} are used to represent the conjunction rule, the conjunction activation function used is as follows:

\begin{equation}
Conj = \prod_{i=1}^{n} 1-w_{i}(1-x_{i}), \;\;\; x_{i},w_{i} \in \left \{ 0,1 \right \} \
\label{equa2}
\end{equation}

Where $w_{i}$ represents the weight, and $x_{i}$ represents the input of the previous layer. It should be noted that both the weight and the input are binary values. Therefore, such a network can be equivalently transformed into a set of conjunction rules. So far, the rules we need can be represented by a neural network. Take fig.~\ref{fig1} as an example, if and only if node 1 and node 2 output one, node 3 will output one.
 
\subsection{Feature Selection and Binarization}

According to suggestions by \cite{ruyu2019comparison}, credit features of the applicant are divided into three categories: loan information, history information, and soft information. Loan information includes features directly related to the loan application. History features contain statistical information about the past behavior of the applicant. Soft information refers to features that are not directly related to lending but may also be helpful for classification.

According to the model design, the input features are to be binarized. Therefore, the original feature values need to be processed by one-hot encoding. One-hot encoding is intuitive for categorical features but not for continuous, binning is therefore required for continuous features. Binning is a common engineering strategy, which introduces non-linear and enhances the robustness of data. Here we use the decision tree algorithm to discretize the continuous feature. In detail, for each continuous feature, the CART is used to train separately, and take the threshold value of the split node as the basis of binning. This is the commonly used binning strategy in credit evaluation modeling. After binning for continuous features, One-hot encoding can be carried out normally.

% 这里缺少one-hot的具体策略 / 加在上面一段了

\subsection{Model Outline and Training}

\begin{figure}
\centering
\includegraphics[width=0.7\textwidth]{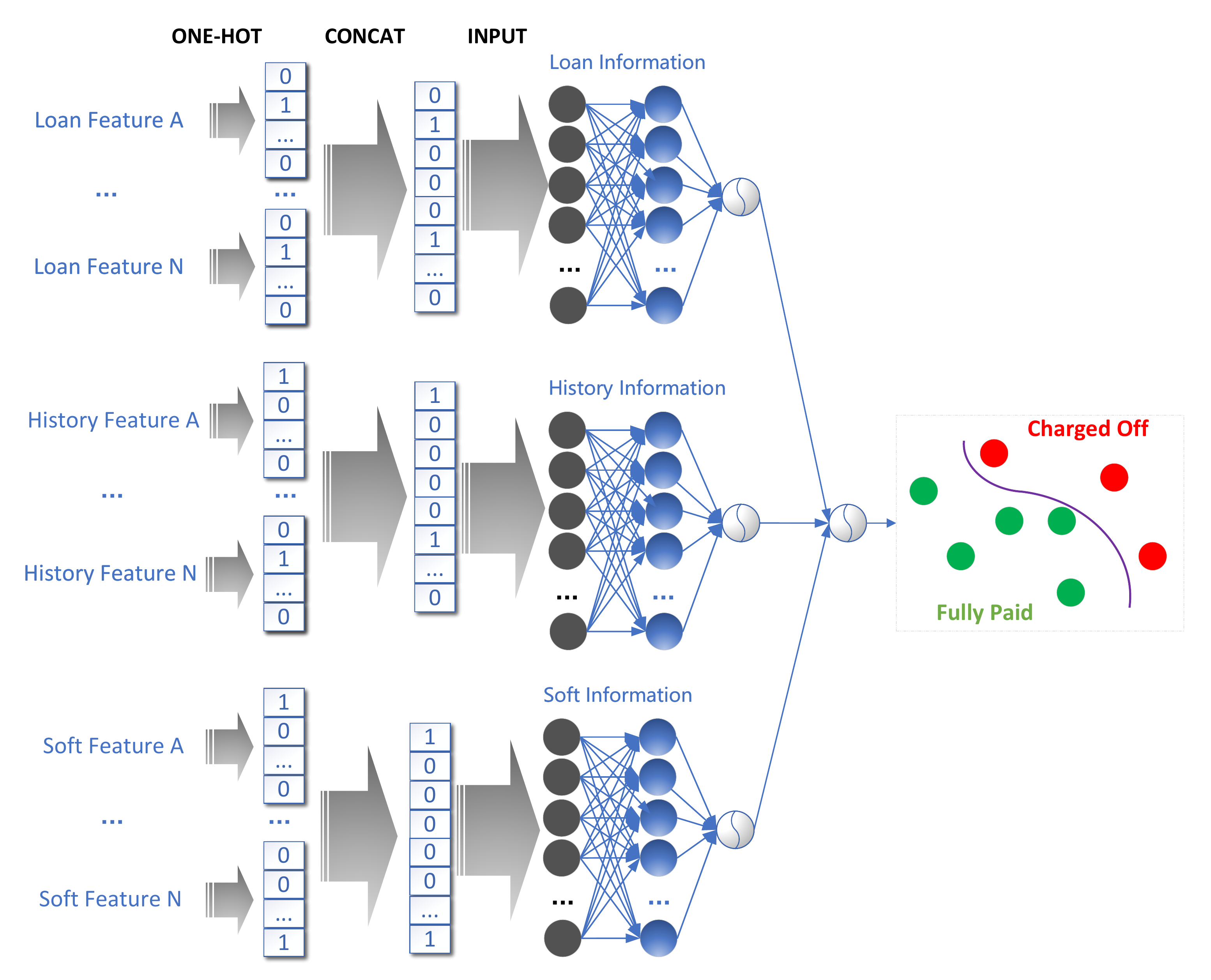}
\caption{Overall modeling structure.} \label{fig2}
\end{figure}

The overall structure consists of three sub-networks and then are aggregated by a fully connected layer to make a final decision. Each sub-network contains only one hidden layer(see Fig.~\ref{fig2}). After one-hot, the input layer (the black nodes) receive the new features after encoding. The hidden layer (the blue nodes) is essentially a fully connected layer, but to simulate the conjunction behaviour between rules, we use the conjunction activation function introduced previously. Each sub-network can be regarded as a sub-classifier, which can output a default probability between 0 and 1 through the sigmoid activation function. So far, the overall network structure has been presented, but there are still problems to be solved in the practical training process.

The most severe obstacle is the vanishing gradient. As mentioned earlier, the weight of the hidden layer should only take 0 or 1. Here, the weight matrix can be understood as an adjacency matrix. Unlike the continuous weight of a regular neural network, this makes training almost impossible. Nevertheless, \cite{wang2021scalable} solved the problem skillfully. They presented a training method for the discrete weight neural network and an improved conjunction activation function to solve the vanishing gradient and the training of large-scale data sets respectively. 

More specifically, We draw lessons from the idea of the Gradient Grafting proposed by \cite{wang2021scalable}. In the training process, discrete and continuous weight models will be maintained simultaneously. The so-called continuous model means that the weights are floating-point numbers between 0 and 1. The discrete model discretizes the hidden layer's weight matrix of the continuous model with a threshold of 0.5. It should be noted that only the hidden layer's weight matrix of the discrete model is binary, and the weights of other layers are the same as those of the continuous model. The weights of the continuous model are clipped between 0 and 1 manually after each update. The regular gradient descent can be seen from Equation \ref{equa3}. However, the update of gradient grafting can be formulated as the Equation \ref{equa4}, where $Y_d$ is the output of the discrete model and $Y_c$ is the output of the continuous model. The updating direction of the weights will be more focused on optimizing the loss of the discrete model. In this way, we effectively train the discrete model actually needed with the help of the continuous one.

\begin{equation}
W_{t+1} = W_{t} - \eta \frac{\partial L(Y_{c})}{\partial W_{t}}
\label{equa3}
\end{equation}

\begin{equation}
W_{t+1} = W_{t} - \eta \frac{\partial L(Y_{d})}{\partial Y_{d}} \cdot \frac{\partial Y_{c}}{\partial W_{t}}
\label{equa4}
\end{equation}

However, when we need to train on a large-scale dataset, we can find that the vanishing gradient is still unavoidable by analyzing the formula of its partial derivative w.r.t. weights. As the Equation \ref{equa5}, both $X_{k} - 1$ and $1-W_{i}(1-X_{i})$ are values between 0 and 1. Multiplication of multiple numbers between 0 and 1 has to cause the entire value to approach 0. Therefore, the improved conjunction activation function proposed by \cite{wang2021scalable} can be used here as the Equation \ref{equa6}. Here, logarithms are used to convert multiplications to additions, and $\frac{-1}{-1+x}$ is aiming to keep the behaviours of the conjunction activation function.

So far, we have achieved efficient training of the discrete neural network by the Gradient Grafting and the improved conjunction activation function. To be precise, we can train each sub-network individually. However, the network will not converge so easily when the three sub-networks are trained together. However, since it can be trained individually, one can first train the three sub-networks separately and then let the trained weights as the initial values of joint training later.

\begin{equation}
\frac{\partial Conj}{\partial W_{k}} = (X_{k} - 1) \cdot \prod_{i=1 \wedge  i \neq k}^{n} 1-W_{i}(1-X_{i})
\label{equa5}
\end{equation}

\begin{equation}
Conj_{+} = \frac{-1}{-1 + log(Conj)}
\label{equa6}
\end{equation}

\section{Experiments}

\subsection{Dataset}

Most of the loan data are out of date and small-scale, not in line with the practical significance. So we take the data from the Lending Club as the experimental data, which is the largest P2P platform in the US. We collected the loan records of the Lending Club from 2007 to the fourth quarter of 2018. Among 1048575 records, we randomly selected 100000 of them for the experiment to save training time. 

The features used and their categories are shown in the table \ref{tab1}. The target variable is loan status, including three values: \emph{Current}, \emph{Fully paid} and \emph{Charged off}. What we actually need here is the record of \emph{Fully paid} and \emph{Charged off}. \emph{Fully paid} refers to the loan has been repaid in full, encoded as 0 in label column, also refers to negative sample in the paper, and \emph{Charged off} encoded as 1, refers to positive sample, they are applicants of loans that have not been repaid within a period of time. So the task can be simplified to a typical binary classification. For features, we refer to the preprocessing methods in \cite{lee2021graph}. However, we do not deal with the data imbalance because we want to avoid the influence of data preprocessing on the performance results as much as possible.

\begin{table}
\centering
\caption{Features selected and categories belong.}\label{tab1}
\begin{tabular}{|c|c|}
\hline
Category & Feature Name \\ \hline
& Installment                  \\
& Loan\,Purpose                \\
& Loan\,Application\,Type      \\
Loan Information    & Interest\,Rate     \\
& Last\,Payment\,Amount        \\
& Loan\,Amount                 \\
& Revolving\,Balance           \\ \hline 
& Delinquency\,In\,2\,Years    \\
& Inquiries\,In\,6\,Months     \\
& Mortgage\,Accounts           \\
History Information & Grade           \\
& Open\,Accounts               \\
& Revolving\,Utilization\,Rate \\
& Total\,Accounts              \\
& Fico\,Avg                    \\ \hline
& Address\,State               \\
& Employment\,Length           \\
Soft Information    & Home\,Ownership     \\
& Verification\,Status         \\
& Annual\,Income               \\ \hline
\end{tabular}
\end{table}

\subsection{Classification Performance}

First of all, it should be noted that performance is not the final goal of our model, but since performance comparison is an issue that must be mentioned, a brief comparison is carried out here.

\subsubsection{CART} The decision tree can also be transformed into a conjunctive rule set similar to our model, so it is reasonable to focus on the performance difference between both. We use the implementation of the decision tree in \cite{scikit-learn} with the default parameters setting.

\subsubsection{Representative Black-box Models} They mainly include some tree-based models, which are usually considered inexplicable due to their complex internal structure, such as random forest, XGBoost, LightGBM, etc. We also use the implementation of these algorithms in \cite{scikit-learn}.

For random forest, we set n\_estimators=100, criterion="gini". For XGBoost, we set learning\_rate=0.01, n\_estimators=160, objective="binary:logistic". As for LightGBM, we keep its default setting. These models are complex enough to be called black box models, because we don't know their decision path at all.

\subsubsection{Our Model} As mentioned before, our model consists of the three sub-networks, each of which has only one hidden layer to more likely preserve the interpretability and understandability of a human. However, the number of nodes in the hidden layer is a hyperparameter to be selected. We tried from 16, 32, 64, 128 to 256, and the learning rate from 0.1, 0.01, 0.001, 0.0001, 0.00001, etc.

Considering that the data set is unbalanced, we will examine the three indicators of Accuracy, F1-Score, and AUC. The results are shown in table \ref{tab2}. It can be seen that on this data, the performance of different models is not much different except for the CART. Due to the need to compromise the interpretability as much as possible in the structure of the model, our model has not surpassed LightGBM and XGBoost in performance. However, it is pretty close. Such loss can be exchanged for structural transparency and interpretability, which is worthwhile for high-stake application scenarios.

Our result also reveals that it is not that more complex models lead to better performance, especially for structured tabular data. There is a myth experience that the complexity of the model is proportionate to the performance. However, such a view should be taken with a grain of salt. Our model are pretty close to tree-based black-box models in performance in practice on this dataset and obviously better than the decision tree with similar structure.

\begin{table}[]
\centering
\caption{Classification performance comparison by 5-fold cross validation, the results have been averaged.}\label{tab2}
\begin{tabular}{|c|cccc|}
\hline
& Model & Accuracy & F1-Score & AUC   \\ \hline
\multicolumn{1}{|c|}{\multirow{2}{*}{Interpretable Models}} & \textbf{Our Model}     & \textbf{0.865} & \textbf{0.655} & \textbf{0.923} \\
\multicolumn{1}{|c|}{}                                      & CART      & 0.822    & 0.570    & 0.731 \\ \hline
\multicolumn{1}{|c|}{\multirow{3}{*}{Black-box Models}}    & \textbf{LightGBM}  & \textbf{0.871} & \textbf{0.665} & \textbf{0.930}  \\
\multicolumn{1}{|c|}{}                                      & XGBoost            & 0.865    & 0.656    & 0.926 \\
\multicolumn{1}{|c|}{}     & Random Forest      & 0.864    & 0.646    & 0.924 \\

 \hline
\end{tabular}
\end{table}

\subsection{Global and Local Explanations}
The model can provide both global and local explanations. Global explanation means that the internal structure of the model is clear and can be equivalently convert into rule sets. It help users predict what kind of behaviour combination will lead to a default. Local interpretation focuses on a specific sample. For example, when an user is rejected, the reasons or factors that lead to the default prediction can be inferred in reverse based on the rule set. The two explanations make the business logic clearer, applicants more responsible and help eliminate discrimination.

\subsubsection{Global Explanation} 

The model can be converted into three rule sets according to the feature classification. As shown in Table \ref{tab3}, the weight of each rule can be obtained after training. The loan information has the highest impact on the prediction, as we expected. For a specific rule, if the weight exceeds 0.5, it represents a negative factor. According to this, a global rule view that reflects positive or negative factors can be drawn. With this global view, experience and knowledge are accumulated to understand and improve the model.

Some rules and common sense confirm each other. For example, 
There may be rules that are in line with our common sense cognition. For example, the rule $(Grade=A, 0.45)$ indicates that credit grade A is a positive factor in avoiding rejection, which matches our prior knowledge. For another example, the rule $((8.0\leq InterestRate\,\textless\,12.0)\wedge(LastPaymentAmount\,\textless\,7.0), 0.52)$ tells us that the smaller the last payment, the negative impact on the prediction, which may be seldom realized in our previous experience, can be extracted as knowledge for domain experts. Some other rules in the model defy common sense. For example, $(Verification = False, 0.49)$ indicates that applicant information verification is a negative factor for loan applications. This is clearly against common sense. Then we have reason to verify the dataset used for training and improve the model. 

In other words, the transparent structure allows us to debug our model efficiently. The simpler the model structure, the faster the iterative upgrading of model performance, rather than the opposite.

% \begin{table}[]
% \centering
% \caption{Rule examples extracted from our model.}\label{tab2}
% \begin{tabular}{|c|c|}
% \hline
% Weight & Rules                              \\ \hline
% 0.42   & Grade = A                          \\ \hline
% 0.53   & Last Payment Amount \textless 7.0  \\ \hline
% 0.47   & Verification Status = Not Verified \\ \hline
% \end{tabular}
% \end{table}

\begin{table}[]
\centering
\caption{Rule examples extracted from our model.}\label{tab3}
\begin{tabular}{|c|c|cc|}
\hline
Subnet Weight            & Subnet                               & \multicolumn{1}{c|}{Weight} & Rules                                                                                                                                                        \\ \hline
\multirow{3}{*}{6.9160}  & \multirow{3}{*}{Loan Information}    & \multicolumn{1}{c|}{0.52}   & \begin{tabular}[c]{@{}c@{}}8.0 \textless{}= InterestRate \textless 12.0\\ AND\\ LastPaymentAmount \textless 7.0\end{tabular}                              \\ \cline{3-4} 
                         &                                      & \multicolumn{1}{c|}{0.43}   & \begin{tabular}[c]{@{}c@{}}LoanApplicationType = Individual\\ AND\\ LastPaymentAmount \textgreater{}= 8.0\\ AND\\ LoanAmount \textless 9.2\end{tabular} \\ \cline{3-4} 
                         &                                      & \multicolumn{2}{c|}{...}                                                                                                                                                                   \\ \hline
\multirow{3}{*}{-0.7029} & \multirow{3}{*}{History Information} & \multicolumn{1}{c|}{0.45}   & Grade = A                                                                                                                                                    \\ \cline{3-4} 
                         &                                      & \multicolumn{1}{c|}{0.50}   & \begin{tabular}[c]{@{}c@{}}MortgageAccounts = 0\\ AND\\ Grade = D or less\end{tabular}                                                                      \\ \cline{3-4} 
                         &                                      & \multicolumn{2}{c|}{...}                                                                                                                                                                   \\ \hline
\multirow{3}{*}{1.0364}  & \multirow{3}{*}{Soft Information}    & \multicolumn{1}{c|}{0.53}   & AnnualIncome \textless 10.9                                                                                                                                 \\ \cline{3-4} 
                         &                                      & \multicolumn{1}{c|}{0.49}   & Verification = False                                                                                                                           \\ \cline{3-4} 
                         &                                      & \multicolumn{2}{c|}{...}                                                                                                                                                                   \\ \hline
\end{tabular}
\end{table}

% \begin{figure}
% \includegraphics[width=\textwidth]{pic/rules_extracted.pdf}
% \caption{Knowledge extracted from model.} \label{fig3}
% \end{figure}

\subsubsection{Local Explanation} 

When someone is rejected by the model, we can provide the applicant with the reasons for the refusal. We only need to check the network nodes activated by the input. From a rule-set perspective, that is, that rule sets that the applicant satisfies. Then show him those combinations of behaviors that have a weight greater than 0.5. In this way, users can be more trusted in the system within the financial institution, and at the same time, they can also guide users' future behavior to get the loan.

% \begin{figure}
% \includegraphics[width=\textwidth]{pic/pic4.pdf}
% \caption{Institutions can justify rejecting a loan application, and applicants can learn how to improve their chances of getting a loan in the future.} \label{fig4}
% \end{figure}

\subsection{Correctness Test of Post-hoc Methods}

The flaws of post-hoc methods were brought to the fore as early as 2019\cite{rudin2019stop}. The explanations provided by the post-hoc method may not be faithful to the actual decision-making behaviour of the original model. However, it is difficult to convince without evidence to support it. 

In this section, our model is used to test the correctness of the explanation generated by post-hoc methods. The basis is that each sub-network in our model can be equivalently converted into a rule set, and its representation form is basically similar to the explanations provided by some post-hoc methods. Anchor\cite{ribeiro2018anchors} is selected as a representative post-hoc method for testing. The method can provide sufficient conditions for model decision results and classify any other samples that meet these conditions as the same class as the explained sample.

Specifically speaking, after building our model, treat it as a black box and then use Anchor to explain positive prediction (\emph{Charged off}) of the model. After that, we can check whether the explanation provided by Anchor is contrary to our actual decision-making paths. For example, the rules provided by Anchor contain features that are not included in our model. Unsurprisingly, we did find such an example.

A sample is classified as charged off by our model, then we can check at the nodes that are activated (the rules satisfied). The decision path of this sample is transparent and formulated as conjunction rules. The Anchor was then used to explain this sample as well. It provides a rule set, precision, and coverage, which means that samples that meet the rule set have a \emph{precision} percent probability of being considered as Charged Off samples by our model. However, we compare rules provided by the Anchor to our actual decision paths, $ loan\_amount \geq 9.5 $  is not in an actual rule set at all, which means that the Anchor provides a wrong interpretation of our model's decision. This case is relatively solid evidence that the post-hoc methods sometimes provide explanations that are not faithful to the behavior of the original model. It is as if we go from one place to another, the original model arrives via path \emph{A}, and the explanation provided by the post-hoc methods arrives via path \emph{B}, both arrive at the same destination, but this is not what we desire.
 
An inaccurate interpretation would give a wrong guide to the users' behavior in the future. For example, the explanation provided by the Anchor contains $ loan\_amount \geq 9.5 $, which means that the loan amount is a significant reason for rejection, which may drive the user to reduce the amount applied for the following loan application. However, this behavior does not actually improve the approval rate of future applications because this rule is not included in the decision path of the original model. Such deceptive explanations may lead users to be more suspicious than trusting decision systems, contrary to the original intention of interpretable machine learning. Obviously, it is not suitable to use the post-hoc methods in high-stake decision-making scenarios.
 
%  \begin{figure}
% \includegraphics[width=\textwidth]{pic/pic6.pdf}
% \caption{The Anchor provides an explanation for infidelity to the original model.} \label{fig5}
% \end{figure}
 
\subsection{Experimental Summary}
From the experimental stage, we first illustrated the results of the classification performance of our model. Results show that the performance of our model is significantly better than that of the decision tree with the most similar structure, which shows that the gradient descent to build a decision tree-like rule set can effectively avoid the local optimum caused by heuristic training. Nevertheless, our performance is not as good as some black-box models, but it is pretty close. The resulting transparency and interpretability are crucial to widespread application in high-stake application scenarios. We pay more attention to the practicality and interpretability of the model, which is more meaningful than improving the performance at the thousandths. The result shows that we got an acceptable trade-off.

Then, we presented the advantages of our model, provide global and local explanations, and analyzed in detail how it breaks the trust barrier between the applicants and financial institutions.

More importantly, we demonstrated the unreliability of post-hoc methods—a practical case of the post-hoc method that are not faithful to the behavior of the original model. Reflected from the side, the correct direction of interpretable machine learning should be considered interpretability before building models rather than post-hoc, especially in application with high risks.

\section{Conclusion}
In FinTech, model interpretability is more of a concern to users than performance. Post-hoc interpretation method cannot accurately explain the internal mechanism of black-box models. RRL represents and learns conjunction logical rules through neural networks, taking into account both interpretability and performance. Based on this advantage, we design decision rules for loans and create interpretable credit evaluation models based on RRL. Experiments demonstrate that the model performs as well as state-of-the-art black-box models while maintaining interpretability, and it has been verified in credit evaluation application scenarios. In addition, using the proposed model as the verification tool, the simulation analysis gives the confirmation that the credibility of the post-hoc methods is insufficient.

There are still many aspects for our work to improve in the future. In the structure of proposed neural network, only the conjunction function is considered. However, the disjunction function is often in loan credit evaluation. The disjunction activation function can be further studied in future research. In otherwise, we can continue to improve the performance, which may be achieved by improving the gradient grafting training trick, the binning of continuous features can be added to the training process of the model. In addition, by communicating with industry practitioners, we can design models that are more in line with the interpretability requirements of credit evaluation or other fields. At a macro level, our efforts should also focus on formal definitions of interpretable machine learning since there has not been a rigorous definition of this topic so far.

\subsubsection{Acknowledgements}  This research is supported by the National Natural Science Foundation of China (Grant No. 62006198).


\bibliographystyle{unsrt}
\bibliographystyle{splncs04}
\bibliography{paper}



% Springer, Heidelberg (2016). \doi{10.10007/123456789_0}



% LNCS Homepage, \url{http://www.springer.com/lncs}. Last accessed 4
% \end{thebibliography}
\end{document}